\pdfoutput=1
\documentclass[11pt,a4paper]{article}
\usepackage[hyperref]{acl2020}
\usepackage{times}
\usepackage{latexsym}

\usepackage{microtype}

\aclfinalcopy 
\def\aclpaperid{1707} 

\usepackage{tikz}
\usetikzlibrary{positioning,arrows.meta,patterns}
\tikzset{execute at begin node=\strut}
\tikzset{center coordinate/.style={
		execute at end picture={\path
			([rotate around={180:#1}]perpendicular cs:
			horizontal line through={#1},
			vertical line through={(current bounding box.east)})
			([rotate around={180:#1}]perpendicular cs:
			horizontal line through={#1},
			vertical line through={(current bounding box.west)});
		}
}}
\tikzstyle{defmeta}=[on grid, style=thick, inner sep=0pt]
\newcommand{\defxy}[3]{\tikzstyle{#1}=[defmeta, node distance=#3 mm and #2 mm, x=#2 mm, y=#3 mm]}
\defxy{default}{25}{21}
\defxy{defsquare}{25}{25}
\defxy{defdeep}{25}{30}
\defxy{defnarrow}{22}{21}
\defxy{defwide}{60}{21}
\defxy{defshort}{25}{15}
\defxy{defcompact}{22}{16}
\tikzstyle{scopetree}=[on grid, node distance = 12mm and 24mm]
\tikzstyle{scopetreedeep}=[on grid, node distance = 18mm and 24mm]
\tikzstyle{node}=[circle, draw, minimum size=10mm]
\tikzstyle{ent}=[node, fill=porange]
\tikzstyle{val}=[node, fill=plila]
\tikzstyle{tok}=[node, fill=grau]
\tikzstyle{sit}=[node, fill=qorange]
\tikzstyle{func}=[node, fill=qlila]
\tikzstyle{bin}=[rectangle, draw, minimum size=2.1mm, inner sep=-2mm, font=\tiny]
\tikzstyle{comp}=[bin, fill=orange]
\tikzstyle{sig}=[bin, fill=lila]
\tikzstyle{link1}=[out=106, in=74, looseness=.5]
\tikzstyle{link2}=[out=102, in=78, looseness=.6]
\tikzstyle{link3}=[out=98, in=82, looseness=.69]
\tikzstyle{link4}=[out=94, in=86, looseness=.77]
\tikzstyle{link5}=[out=90, in=90, looseness=.84]
\tikzstyle{dir}=[-{Straight Barb[angle=60:2.2mm]}, shorten > =0.2mm]
\tikzstyle{dircomp}=[{Straight Barb[angle=90:2.2mm]}-, shorten < =0.2mm]
\tikzstyle{dirextra}=[shorten < =0.5mm]
\tikzstyle{inf}=[densely dotted, {Straight Barb[angle=60:2.2mm]}-, shorten < =0.2mm]
\tikzstyle{scope}=[dotted, lila, {Straight Barb[angle=60:2.2mm]}-, shorten < =0.2mm]
\newlength{\compsep}
\newlength{\scopesep}
\newlength{\funcsep}
\definecolor{lila}{RGB}{159,114,207}
\definecolor{orange}{RGB}{230,130,50}
\colorlet{plila}{lila!60!white}
\colorlet{qlila}{lila!20!white}
\colorlet{porange}{orange!60!white}
\colorlet{qorange}{orange!20!white}
\definecolor{braun}{tHsb}{27,1,.5}
\definecolor{violett}{tHsb}{269,1,.5}
\definecolor{hellviolett}{tHsb}{269,1,.8}
\definecolor{grau}{RGB}{192,192,192}

\usepackage{amssymb}
\usepackage{amsmath}	
\usepackage{nicefrac}
\usepackage{xfrac}
\usepackage{bbm}

\newcommand{\eq}{\!=\!}  
\newcommand{\mn}{\!-\!}

\newcommand{\rin}{\;\mathrm{in}\;}  
\newcommand{\psp}{^{\vphantom{()}}}  
\newcommand{\rd}[1]{\left(#1\right)}  
\newcommand{\sq}[1]{\left[#1\right]}  
\newcommand{\cl}[1]{\left\{#1\right\}}  
\newcommand{\del}[1]{\frac{\partial}{\partial #1}}  
\newcommand{\dth}{\del{\theta}}  
\newcommand{\dthb}[1]{\dth\rd{#1\big.}}  
\newcommand{\ene}[1]{\exp\rd{-E(#1)\big.}}  

\newcommand{\E}[1]{\mathbb{E}_{#1}}  
\newcommand{\Eo}[2]{\E{#1}\sq{#2}}
\renewcommand{\Pr}{\mathbb{P}}  
\newcommand{\Po}[1]{\Pr\rd{#1}}  
\newcommand{\Pc}[2]{\Po{#1\,\middle|\,#2}}  
\newcommand{\Q}{\mathbb{Q}}  

\usepackage[nameinlink,capitalize]{cleveref}  
\crefname{equation}{}{}
\crefformat{section}{#2\S#1#3}
\crefformat{subsection}{#2\S#1#3}
\crefmultiformat{section}{#2\S#1#3}{ and~#2\S#1#3}{, #2\S#1#3}{, and~#2\S#1#3}
\crefmultiformat{subsection}{#2\S#1#3}{ and~#2\S#1#3}{, #2\S#1#3}{, and~#2\S#1#3}
\crefrangeformat{section}{#3\S#1#4--#5#2#6}
\crefrangeformat{subsection}{#3\S#1#4--#5#2#6}

\newcommand{\citex}[3][]{\citep[#2:][#1]{#3}}  
\newcommand{\citeg}[2][]{\citex[#1]{for example}{#2}}  

\newcommand{\sortbib}[1]{}

\usepackage{multirow}

\title{Autoencoding Pixies: Amortised Variational Inference
with Graph Convolutions for Functional Distributional Semantics}

\author{Guy Emerson \\
  Department of Computer Science and Technology \\
  University of Cambridge \\
  \texttt{gete2@cam.ac.uk}}

\date{}

\begin{document}
\maketitle
\begin{abstract}
Functional Distributional Semantics
provides a linguistically interpretable framework for distributional semantics,
by representing the meaning of a word as a function (a binary classifier),
instead of a vector.
However, the large number of latent variables
means that inference is computationally expensive,
and training a model is therefore slow to converge.
In this paper, I introduce the Pixie Autoencoder,
which augments the generative model of Functional Distributional Semantics
with a graph-convolutional neural network
to perform amortised variational inference.
This allows the model to be trained more effectively,
achieving better results on two tasks
(semantic similarity in context and semantic composition),
and outperforming BERT, a large pre-trained language model.
\end{abstract}

\section{Introduction}

The aim of distributional semantics is to learn the meanings of words from a corpus
\citep{harris1954distribution,firth1951collocation,firth1957company}.
Many approaches learn a vector for each word,
including count models and embedding models
\citex{for an overview, see}{erk2012vector,clark2015vector},
and some recent approaches learn a vector for each token in a particular context
\citeg{peters2018elmo,devlin2019bert}.

However, such vector representations do not
make a clear distinction between words and the things they refer to.
This means that such models are challenging to interpret semantically.
In contrast, Functional Distributional Semantics \citep{emerson2016}
aims to provide a framework which can be interpreted in terms of model theory,
a standard approach to formal semantics.

Furthermore, this framework supports first-order logic,
where quantifying over logical variables
is replaced by marginalising out random variables
\citep{emerson2017b,emerson2020quant}.
This connection to logic
is a clear strength over vector-based models.
Even the linguistically inspired tensor-based framework of
\citet{coecke2010tensor} and \citet{baroni2014tensor}
cannot model quantifiers, as shown by \citet{grefenstette2013tensor}.

However, the linguistic interpretability of Functional Distributional Semantics
comes at a computational cost,
with a high-dimensional latent variable for each token.
Training a model by gradient descent requires
performing Bayesian inference over these latent variables,
which is intractable to calculate exactly.
The main theoretical contribution of this paper
is to present an amortised variational inference algorithm
to infer these latent variables.
This is done using a graph-convolutional network,
as described in~\cref{sec:approach}.

The main empirical contribution of this paper
is to demonstrate that the resulting system,
the Pixie Autoencoder,
improves performance on two semantic tasks,
as described in~\cref{sec:eval}.
I also present the first published results of applying
a large language model (BERT) to these tasks,
showing that results are sensitive to linguistic detail
in how the model is applied.
Despite being a smaller model trained on less data,
the Pixie Autoencoder outperforms BERT on both tasks.

While the proposed inference network is designed for Functional Distributional Semantics,
the proposed techniques should also be of wider interest.
From a machine learning perspective,
amortised variational inference with graph convolutions (\cref{sec:approach:autoencode})
could be useful in other tasks where the input data is a graph,
and the use of belief propagation to reduce variance (\cref{sec:approach:bp})
could be useful for training other generative models.
However, the most important contribution of this work
is from a computational semantics perspective.
This paper takes an important step towards truth-conditional distributional semantics,
showing that truth-conditional functions can be efficiently learnt from a corpus.

\section{Functional Distributional Semantics}
\label{sec:back}

\begin{figure}
	\centering
	\scriptsize
	\begin{tikzpicture}[on grid, very thick, x=6mm, y=5.5mm]
	\draw (-0.2,-0.2) rectangle (12, 5.2);
	\path (1,1) node {\includegraphics[width=5mm]{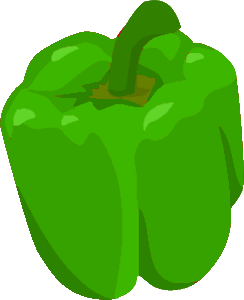}} ++(0.55,-0.4) node {1};
	\path (2.2,3) node {\includegraphics[width=5mm]{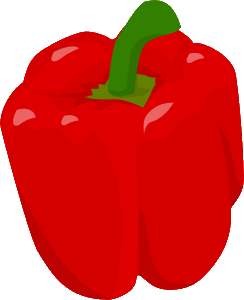}} ++(0.55,-0.4) node {5};
	\path (1,4) node {\includegraphics[width=5mm]{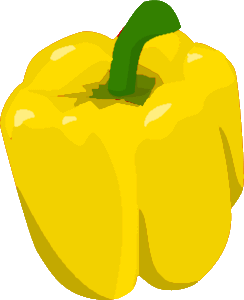}} ++(0.55,-0.4) node {2};
	\path (3,1.5) node {\includegraphics[width=5mm]{green-pepper}} ++(0.6,-0.4) node {14};
	\path (3.5,3.5) node {\includegraphics[width=5mm]{green-pepper}} ++(0.6,-0.4) node {12};
	\path (4.5,1) node {\includegraphics[width=5mm]{yellow-pepper}} ++(0.55,-0.4) node {3};
	\path (5,2.5) node {\includegraphics[width=5mm]{red-pepper}} ++(0.55,-0.4) node {8};
	\path (6.2,3.7) node {\includegraphics[width=5mm]{red-pepper}} ++(0.55,-0.4) node {7};
	\path (6.7,2) node {\includegraphics[width=5mm]{yellow-pepper}} ++(0.6,-0.4) node {11};
	\path (8.5,2.5) node {\includegraphics[width=5mm]{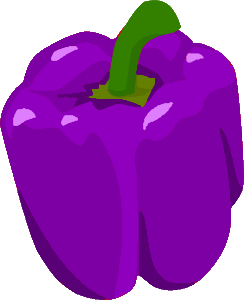}} ++(0.55,-0.4) node {4};
	\path (11,3) node {\includegraphics[width=5mm]{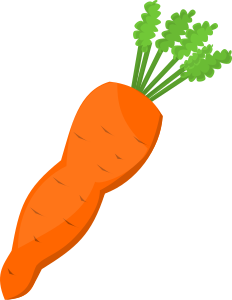}} ++(0.15,-0.4) node {6};
	\path (10.3,4) node {\includegraphics[width=5mm]{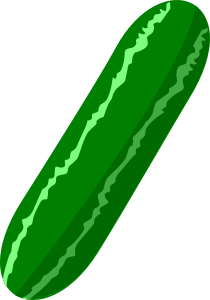}} ++(0.3,-0.4) node {10};
	\path (9.9,1) node {\includegraphics[width=5mm]{cucumber}} ++(0.25,-0.4) node {9};
	\path (11.2,1.5) node {\includegraphics[width=5mm]{carrot}} ++(0.15,-0.4) node {13};
	\draw[orange] (7,0) -- (0,0) -- (0,5) -- (7,5);
	\draw[orange] (7,0) arc [start angle=-90, end angle=90, radius=13.75mm];
	\end{tikzpicture}
	\caption{%
		An example model structure with 14 individuals.
		Subscripts distinguish individuals with identical features,
		but are otherwise arbitrary.
		The \textit{pepper} predicate is true of individuals inside the orange line,
		but the positions of individuals are otherwise arbitrary.
	}
	\vspace*{-3mm}
	\label{fig:pepper-model}
\end{figure}

In this section, I summarise previous work on Functional Distributional Semantics.
I begin in~\cref{sec:back:model} by introducing model-theoretic semantics,
which motivates the form of the machine learning model.
I then explain in~\cref{sec:back:sem-func}
how the meaning of a word is represented as a binary classifier,
and finally present the probabilistic graphical model in~\cref{sec:back:prob-graph}.

\subsection{Model-Theoretic Semantics}
\label{sec:back:model}

The basic idea of model-theoretic semantics
is to define meaning in terms of \textit{truth},
relative to \textit{model structures}.
A model structure can be understood as a model of the world.
In the simplest case, it consists of a set of \textit{individuals}
(also called \textit{entities}),
as illustrated in \cref{fig:pepper-model}.
The meaning of a content word is called a \textit{predicate},
and is formalised as a \textit{truth-conditional function},
which maps individuals to \textit{truth values}
(either \textit{truth} or \textit{falsehood}).

Because of this precisely defined notion of truth,
model theory naturally supports logic,
and has become a prominent approach to formal semantics.
For example, if we know the truth-conditional functions
for \textit{pepper} and \textit{red},
we can use first-order logic to calculate the truth of sentences like
\textit{Some peppers are red},
for model structures like \cref{fig:pepper-model}.

For detailed expositions, see:
\citet{cann1993sem,allan2001sem,kamp2013sem}.

\subsection{Semantic Functions}
\label{sec:back:sem-func}

Functional Distributional Semantics \citep{emerson2016,emerson2018}
embeds model-theoretic semantics into a machine learning model.
An individual is represented by a feature vector, called a \textit{pixie}.\footnote{%
	Terminology introduced by \citet{emerson2017a}.
	This provides a useful shorthand
	for ``feature representation of an individual''.
}
For example, all three red pepper individuals in \cref{fig:pepper-model}
would be represented by the same pixie,
as they have the same features.
A predicate is represented by a \textit{semantic function},
which maps pixies to probabilities of truth.
For example, the function for \textit{pepper}
should map the red pepper pixie to a probability close to~1.
This can be seen in formal semantics
as a truth-conditional function,
and in a machine learning
as a binary classifier.

This ties in with a view of concepts as abilities,
as proposed in some schools of philosophy \citeg{dummett1976meaning,dummett1978know,kenny2010concept,sutton2015prob,sutton2017prob},
and some schools of cognitive science 
(for example:
\citealp{labov1973cup,mccloskey1978judge};
\citealp[pp.~1--3, 134--138]{murphy2002concept};
\citealp{zentall2002concept}).
In NLP, some authors have suggested representing concepts as classifiers,
including \citet{larsson2013classifier},
working in the framework of Type Theory with Records \citep{cooper2005type,cooper2015prob}.
Similarly, \citet{schlangen2016classifier} and
\citet{zarriess2017classifier,zarriess2017classifier2}
train image classifiers using captioned images.

We can also view such a classifier as defining
a region in the space, as argued for by \citet{gaerdenfors2000space,gaerdenfors2014space}.
This idea is used for distributional semantics by \citet{erk2009region1,erk2009region2},
for colour terms by \citet{mcmahan2015colour},
and for knowledge base completion by \citet{bouraoui2017space}.

For a broader survey motivating the use of classifiers to represent meaning,
see: \citet{emerson2020goals}.

\subsection{Probabilistic Graphical Model}
\label{sec:back:prob-graph}

\begin{figure}[t]
	\centering%
	\vspace*{2mm}
	\begin{tikzpicture}[default]
	\tikzstyle{word}=[inner sep=1mm]
	
	\draw[white] (-0.5, 0) -- (2.5, 0) ; 
	
	\node[word] (x) {picture};
	\node[word, right=of x] (y) {tell};
	\node[word, right=of y] (z) {story};
	
	\draw[dir] (y) -- (x) node[midway, above, xshift=1mm] {\textsc{arg1}} ;
	\draw[dir] (y) -- (z) node[midway, above, xshift=-1mm] {\textsc{arg2}} ;
	
	\end{tikzpicture}%
	\vspace*{-1mm}%
	\caption[]{%
		A dependency graph,
		which could be generated by \cref{fig:graph}.
		Such graphs are observed in training.
	}%
	\label{fig:dmrs}%
\end{figure}
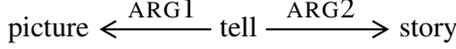

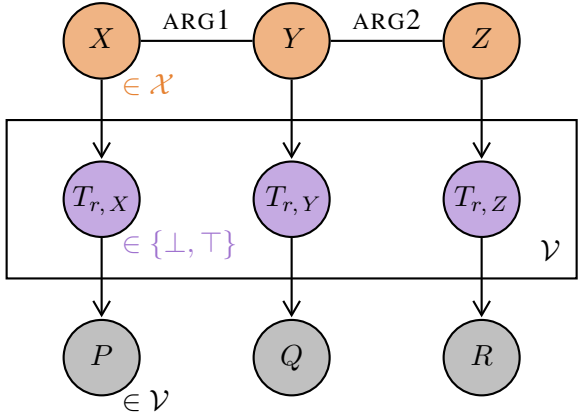
\begin{figure}[t]%
	\centering%
	\begin{tikzpicture}[default]
	
	
	\node[ent] (y) {$Y\!$} ;
	\node[ent, right=of y] (z) {$Z$} ;
	\node[ent, left=of y] (x) {$X$} ;
	
	\draw (y) -- (z) node[midway, above, color=black] {\textsc{arg2}} ;
	\draw (y) -- (x) node[midway, above, color=black] {\textsc{arg1}} ;
	
	\node[below right=3.5mm and 2.5mm of x, anchor=north west] {\textcolor{orange}{$\in \mathcal{X}$}} ;
	
	
	\draw (-1.5,-0.5) rectangle (1.5,-1.5) ;
	
	\node[val, below=of x] (tx) {$T_{r,\,X}$} ;
	\node[val, below=of y] (ty) {$T_{r,\,Y\!}$} ;
	\node[val, below=of z] (tz) {$T_{r,\,Z}$} ;
	
	\draw[dir] (x) -- (tx);
	\draw[dir] (y) -- (ty);
	\draw[dir] (z) -- (tz);
	
	\node[below right=3.5mm and 2.5mm of tx, anchor=north west] {\textcolor{lila}{$\in \cl{\bot,\top}$}} ;
	\node[xshift=-2ex, yshift=2ex] at (1.5, -1.5) {$\mathcal{V}$};
	
	
	\node[tok, below=of tx] (a) {$P$} ;
	\node[tok, below=of ty] (b) {$Q$} ;
	\node[tok, below=of tz] (c) {$R$} ;
	
	\draw[dir] (tx) -- (a);
	\draw[dir] (ty) -- (b);
	\draw[dir] (tz) -- (c);
	
	\node[below right=3.5mm and 2.5mm of a, anchor=north west] {$\in \mathcal{V}$} ;
	
	\end{tikzpicture}%
	\vspace*{-1mm}
	\caption{%
		Probabilistic graphical model for Functional Distributional Semantics.
		Each node is a random variable.
		The plate (box in middle) denotes repeated nodes.
		\newline
		\textbf{Top row:}
		individuals represented by jointly distributed
		pixie-valued random variables $X$, $Y$,~$Z$,
		in a space~$\mathcal{X}$.
		This is modelled by a Cardinality Restricted Boltzmann Machine (CaRBM),
		matching the graph topology.
		\newline
		\textbf{Middle row:}
		for each individual, each predicate~$r$ in the vocabulary~$\mathcal{V}$
		is randomly true ($\top$) or false ($\bot$),
		according to the predicate's semantic function.
		Each function is modelled by a feedforward neural net.
		\newline
		\textbf{Bottom row:}
		for each individual, we randomly generate one predicate,
		out of all predicates true of the individual.
		Only these nodes are observed.
	}%
	\label{fig:graph}%
\end{figure}

To learn semantic functions in distributional semantics,
\citeauthor{emerson2016} define a probabilistic graphical model
that generates semantic dependency graphs,
shown in \cref{fig:graph}.
The basic idea is that an observed dependency graph
is true of some unobserved situation
comprising a number of individuals.
Given a sembank (a corpus parsed into dependency graphs),
the model can be trained unsupervised,
to maximise the likelihood of generating the data.
An example graph is shown in \cref{fig:dmrs},
which corresponds to sentences like
\textit{Every picture tells a story}
or \textit{The story was told by a picture}
(note that only content words have nodes).

More precisely, given a \textit{graph topology}
(a dependency graph where the edges are labelled but the nodes are not),
the model generates a predicate for each node.
Rather than directly generating predicates,
the model assumes that each predicate
describes an unobserved individual.\footnote{%
	This assumes a neo-Davidsonian approach to event semantics
	\citep{davidson1967event,parsons1990event},
	where verbal predicates are true of event individuals.
	It also assumes that a plural noun corresponds to a plural individual,
	which would be compatible with \citet{link1983plural}'s approach to plural semantics.
}
The model first generates a pixie to represent each individual,
then generates a truth value for each individual and each predicate in the vocabulary,
and finally generates a single predicate for each individual.
The pixies and truth values can be seen as a probabilistic model structure,
which supports a probabilistic first-order logic
\citep{emerson2017b,emerson2020quant}.
This is an important advantage over other approaches to distributional semantics.

A pixie is defined to be a sparse binary-valued vector,
with $D$~\textit{units} (dimensions),
of which exactly~$C$ are \textit{active} (take the value~1).\footnote{%
	Although a pixie is a feature vector,
	the features are all latent	in distributional semantics,
	in common with models like
	LDA \citep{blei2003lda} or Skip-gram \citep{mikolov2013vector}.
}
The joint distribution over pixies
is defined by a Cardinality Restricted Boltzmann Machine (CaRBM) \citep{swersky2012carbm},
which controls how the active units of each pixie should co-occur
with the active units of other pixies in the same dependency graph.

A CaRBM is an energy-based model,
meaning that the probability of a situation
is proportional to the exponential of the negative \textit{energy} of the situation.
This is shown in~\cref{eqn:sit-dist-ene},
where $s$~denotes a \textit{situation} comprising
a set of pixies with semantic dependencies between them,
and $E(s)$~denotes the energy.
The energy is defined in~\cref{eqn:sit-dist},\footnote{%
	I follow the Einstein summation convention,
	where a repeated subscript is assumed to be summed over.
	For example, $x_i y_i$ is a dot product.
	Furthermore, I use uppercase for random variables,
	and lowercase for values.
	I abbreviate $\Po{X\eq x}$ as $\Po{x}$,
	and I abbreviate $\Po{T_{r,X}\eq\top}$ as $\Po{t_{r,X}}$.
}
where ${x\xrightarrow{l} y}$ denotes
a dependency from pixie~$x$ to pixie~$y$ with label~$l$.
The CaRBM includes a weight matrix $w^{(l)}$ for each label~$l$.
The entry $w^{(l)}_{ij}$ controls how likely it is for
units $i$ and~$j$ to both be active,
when linked by dependency~$l$.
Each graph topology has a corresponding CaRBM,
but the weight matrices are shared across graph topologies.
Normalising the distribution in~\cref{eqn:sit-dist}
is intractable, as it requires summing over all possible~$s$.
\vspace*{-1mm}
\begin{align}
\Po{s}
&\propto \ene{s}
\label{eqn:sit-dist-ene}
\\
\Po{s}
&\propto \exp\rd{
	\sum_{x\xrightarrow{l} y \rin s}\!\!\!
	w^{(l)}_{ij} x\psp_i y\psp_j
}
\label{eqn:sit-dist}
\end{align}
\vspace*{-2mm}

The semantic function~$t^{(r)}$ for a predicate~$r$
is defined to be one-layer feedforward net,
as shown in~\cref{eqn:sem-func},
where $\sigma$ denotes the sigmoid function.
Each predicate has a vector of weights~$v^{(r)}$.

\vspace*{-2mm}
\begin{equation}
t^{(r)}(x)
= \sigma\rd{
	v^{(r)}_i x\psp_i
}
\label{eqn:sem-func}
\end{equation}

Lastly, the probability of generating a predicate~$r$ for a pixie~$x$
is given in~\cref{eqn:gen-pred}.
The more likely $r$~is to be true,
the more likely it is to be generated.
Normalising requires summing over the vocabulary.
\vspace*{-2mm}
\begin{equation}
\Pc{r}{x}
\propto t^{(r)}(x)
\label{eqn:gen-pred}
\end{equation}

In summary, the model has parameters $w^{(l)}$ (the world model),
and $v^{(r)}$ (the lexical model).
These are trained on a sembank using the gradients in~\cref{eqn:grad},
where $g$~is a dependency graph.
For~$w^{(l)}$, only the first term is nonzero;
for~$v^{(r)}$, only the second term.
\vspace*{-5mm}
\begin{equation}
\begin{split}
\dth \log\Po{g}
=& \rd{\Big. \E{s|g} - \E{s} } \sq{\dthb{-E(s)} } \\
& + \E{s|g} \sq{\dth\log\Pc{g}{s}}
\end{split}
\label{eqn:grad}
\end{equation}

\section{The Pixie Autoencoder}
\label{sec:approach}

A practical challenge for Functional Distributional Semantics
is training a model in the presence of
high-dimensional latent variables.
In this section, I present the Pixie Autoencoder,
which augments the generative model
with an encoder that predicts these latent variables.

For example, consider dependency graphs for
\textit{The child cut the cake}
and \textit{The gardener cut the grass}.
These are true of rather different situations.
Although the same verb is used in each,
the pixie for \textit{cut} should be different,
because they describe events with different physical actions and different tools
(slicing with a knife vs. driving a lawnmower).
Training requires inferring posterior distributions for these pixies,
but exact inference is intractable.

In \cref{sec:approach:amortise,sec:approach:convolve}, I describe previous work:
amortised variational inference is useful to efficiently predict latent variables;
graph convolutions are useful when the input is a graph.
In~\cref{sec:approach:autoencode}, I present the encoder network,
to predict latent pixies in Functional Distributional Semantics.
It uses the tools introduced in \cref{sec:approach:amortise,sec:approach:convolve},
but modified to better suit the task.
In~\cref{sec:approach:bp}, I explain how the encoder network
can be used to train the generative model,
since training requires the latent variables.
Finally, I summarise the architecture in~\cref{sec:approach:summary},
and compare it to other autoencoders in~\cref{sec:approach:prior}.

\subsection{Amortised Variational Inference}
\label{sec:approach:amortise}
\vspace*{-.5mm}

Calculating the gradients in~\cref{eqn:grad}
requires taking expectations over situations
(both the marginal expectation~$\E{s}$,
and the conditional expectation~$\E{s|g}$ given a graph).
Exact inference would require summing over all possible situations,
which is intractable for a high-dimensional space.

This is a general problem when working with probabilistic models.
Given an intractable distribution~$\Pr(x)$,
a \textit{variational inference} algorithm
approximates this by a simpler distribution~$\Q(x)$,
parametrised by~$q$,
and then optimises the parameters so that $\Q$~is as close as possible to~$\Pr$,
where closeness is defined using KL-divergence
\citex{for a detailed introduction, see}{jordan1999variational}.

However, variational inference algorithms
typically require many update steps in order to
optimise the approximating distribution~$\Q$.
An \textit{amortised variational inference} algorithm
makes a further approximation,
by estimating the parameters~$q$ using an inference network
\citep{kingma2013vae,rezende2014vae,titsias2014vae}.
The inference network might not predict the optimal parameters,
but the calculation can be performed efficiently,
rather than requiring many update steps.
The network has its own parameters~$\phi$,
which are optimised so that it makes good predictions
for the variational parameters~$q$.

\subsection{Graph Convolutions}
\label{sec:approach:convolve}

For graph-structured input data,
a standard feedforward neural net is not suitable.
In order to share parameters across similar graph topologies,
an appropriate architecture is a \textit{graph-convolutional} network
\citep{duvenaud2015graph,kearnes2016graph,kipf2017graph,gilmer2017graph}.
This produces a vector representation for each node in the graph,
calculated through a number of layers.
The vector for a node in layer~$k$
is calculated based only on the vectors in layer~$k{-}1$ for that node
and the nodes connected to it.
The same weights are used for every node in the graph,
allowing the network to be applied to different graph topologies.

For linguistic dependency graphs,
the dependency labels carry important information.
\citet{marcheggiani2017graph} propose
using a different weight matrix for each label in each direction.
This is shown in~\cref{eqn:conv},
where: $h^{(k,X)}$~denotes the vector representation of node~$X$ in layer~$k$;
$w^{(k,l)}$~denotes the weight matrix for dependency label~$l$ in layer~$k$;
$f$~is a non-linear activation function;
and the sums are over outgoing and incoming dependencies.\footnote{%
	For consistency with~\cref{fig:graph},
	I write $X$ for a node (a random variable),
	rather than $x$ (a pixie).
}
There is a separate weight matrix~$w^{(k,l^{-1})}$
for a dependency in the opposite direction,
and as well as a matrix~$w^{(k,\textrm{self})}$
for updating a node based on itself.
Bias terms are not shown.
\vspace*{-1mm}
\begin{equation}
\begin{split}
h^{(k,X)}_i = f\bigg(
& w^{(k,\textrm{self})}_{ij} h^{(k-1,X)}_j \\
& + \sum_{Y\xleftarrow{l}X} w^{(k,l)}_{ij} h^{(k-1,Y)}_j \\
& + \sum_{Y\xrightarrow{l}X} w^{(k,l^{-1})}_{ij} h^{(k-1,Y)}_j \bigg)
\end{split}
\label{eqn:conv}
\end{equation}

\vspace*{-1mm}
\subsection{Predicting Pixies}
\label{sec:approach:autoencode}

\begin{figure}
	\centering
	\begin{tikzpicture}[defdeep]
	
	
	\node[ent] (y) {$Y\!$} ;
	\node[ent, right=of y] (z) {$Z$} ;
	\node[ent, left=of y] (x) {$X$} ;
	
	
	\node[node, below=of x] (tx) {$h^{(X)}$} ;
	\node[node, below=of y] (ty) {$h^{(Y)}$} ;
	\node[node, below=of z] (tz) {$h^{(Z)}$} ;
	
	
	\node[tok, below=of tx] (p) {$e^{(p)}$} ;
	\node[tok, below=of ty] (q) {$e^{(q)}$} ;
	\node[tok, below=of tz] (r) {$e^{(r)}$} ;
	
	
	\draw[inf] (tx) -- (p) node[midway,above,sloped,rotate=180] {\small 1,self};
	\draw[inf] (ty) -- (p) node[near end,above,sloped] {\small 1,\textsc{arg1}};
	\draw[inf] (tx) -- (q) node[near end,above,sloped] {\small 1,\textsc{arg1}$^{-1}$};
	\draw[inf] (ty) -- (q) node[midway,above,sloped,rotate=180] {\small 1,self};
	\draw[inf] (tz) -- (q) node[near end,above,sloped] {\small 1,\textsc{arg2}$^{-1}$};
	\draw[inf] (ty) -- (r) node[near end,above,sloped] {\small 1,\textsc{arg2}};
	\draw[inf] (tz) -- (r) node[midway,above,sloped,rotate=180] {\small 1,self};
	\draw[inf] (x) -- (tx) node[midway,above,sloped,rotate=180] {\small 2,self};
	\draw[inf] (y) -- (tx) node[near end,above,sloped] {\small 2,\textsc{arg1}};
	\draw[inf] (x) -- (ty) node[near end,above,sloped] {\small 2,\textsc{arg1}$^{-1}$};
	\draw[inf] (y) -- (ty) node[midway,above,sloped,rotate=180] {\small 2,self};
	\draw[inf] (z) -- (ty) node[near end,above,sloped] {\small 2,\textsc{arg2}$^{-1}$};
	\draw[inf] (y) -- (tz) node[near end,above,sloped] {\small 2,\textsc{arg2}};
	\draw[inf] (z) -- (tz) node[midway,above,sloped,rotate=180] {\small 2,self};
	
	\end{tikzpicture}%
	\caption{%
		Graph-convolutional inference network for \cref{fig:graph}.
		The aim is to predict the posterior distribution
		over the pixie nodes $X$, $Y$,~$Z$,
		given the observed predicates $p$, $q$,~$r$.
		Each edge indicates the weight matrix used in the graph convolution,
		as defined in~\cref{eqn:conv}.
		In the bottom row, the input at each node
		is an embedding for the node's predicate.
		The intermediate representations~$h$
		do not directly correspond to any random variables in \cref{fig:graph}.
		Conversely, the truth-valued random variables in \cref{fig:graph}
		are not directly represented here.
	}%
	\label{fig:encoder}%
	\vspace*{-.5mm}%
\end{figure}
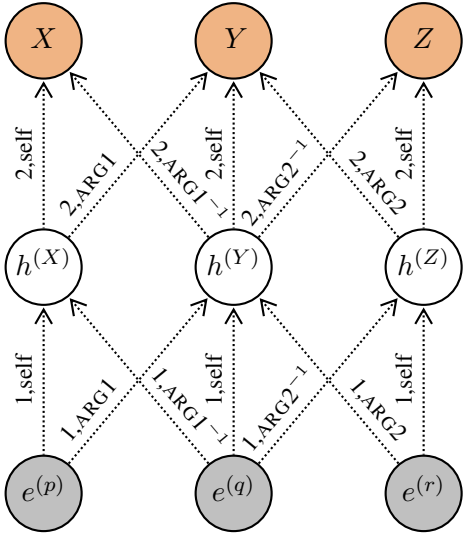

For Functional Distributional Semantics,
\citet{emerson2017a} propose a \textit{mean-field} variational inference algorithm,
where $\Q$ has an independent probability~$q^{(X)}_i$
of each unit~$i$ being active, for each node~$X$.
Each probability is optimised based on the mean activation of all other units.

This makes the simplifying assumption
that the posterior distribution can be approximated
as a single situation with some uncertainty in each dimension.
For example, for a dependency graph for \textit{The gardener cut the grass},
three mean vectors are inferred,
for the gardener, the cutting event, and the grass.
These vectors are ``contextualised'',
because they are jointly inferred based on the whole graph.

I propose using a graph-convolutional network
to amortise the inference of the variational mean-field vectors.
In particular, I use the formulation in~\cref{eqn:conv}, with two layers.
The first layer has a tanh activation,
and the second layer has a sigmoid (to output probabilities).
In addition, if the total activation in the second layer
is above the total cardinality~$C$,
the activations are normalised to sum to~$C$.
The network architecture is illustrated in \cref{fig:encoder}.

The network is trained to minimise the KL-divergence
from $\Pc{s}{g}$ (defined by the generative model)
to $\Q(s)$ (defined by network's output).
This is shown in~\cref{eqn:kl},
where $\E{\Q(s)}$ denotes an expectation over $s$ under the variational distribution.
\vspace*{-3.5mm}
\begin{equation}
D(\Q\|\Pr) = -\Eo{\Q(s)}{\log \rd{\frac{\Pc{s}{g}}{\Q(s)}}}
\label{eqn:kl}
\end{equation}

\vspace*{1mm}
To minimise the KL-divergence,
we can differentiate with respect to the inference network parameters~$\phi$.
This gives~\cref{eqn:grad-phi},
where $H$ denotes entropy.
\vspace*{-4mm}
\begin{equation}
\begin{split}
\del{\phi}D(\Q\|\Pr)
={}&- \del{\phi} \Eo{\Q(s)}{\log\Pr(s)\big.} \\
&- \del{\phi} \Eo{\Q(s)}{\log\Pc{g}{s}\big.} \\
&- \del{\phi} H(\Q)
\label{eqn:grad-phi}
\end{split}
\end{equation}

The first term can be calculated exactly,
because the log probability is proportional to the negative energy,
which is a linear function of each pixie,
and the normalisation constant is independent of $s$ and~$\Q$.
This term therefore simplifies to
the energy of the mean-field pixies, ${\del{\phi}E\rd{\Eo{}{s}}}$.

The last term can be calculated exactly,
because $\Q$~was chosen to be simple.
Since each dimension is independent,
it is ${\sum_q q\log q + (1\mn q)\log (1\mn q)}$,
summing over the variational parameters.

The second term is more difficult, for two reasons.
Firstly, calculating the probability of generating a predicate
requires summing over all predicates,
which is computationally expensive.
We can instead sum over a random sample of predicates
(along with the observed predicate).
However, by ignoring most of the vocabulary,
this will overestimate the probability of generating the correct predicate.
I have mitigated this by upweighting this term,
similarly to a $\beta$-VAE \citep{higgins2017betavae}.

The second problem is that the log probability of a predicate being true
is not a linear function of the pixie.
The first-order approximation would be
to apply the semantic function to the mean-field pixie,
as suggested by \citet{emerson2017a}.
However, this is a poor approximation
when the distribution over pixies has high variance.
By approximating a sigmoid using a probit
and assuming the input is approximately Gaussian,
we can derive~\cref{eqn:probit} \citep[\S8.4.4.2]{murphy2012textbook}.
Intuitively, the higher the variance,
the closer the expected value to~$\nicefrac{1}{2}$.
For a Bernoulli distribution with probability~$q$,
scaled by a weight~$v$, the variance is~$v^2 q(1\mn q)$.
\vspace*{-2mm}
\begin{equation}
\Eo{}{\sigma\!\rd{x}}
\approx
\sigma\!\rd{\frac{
	\E{}\!\sq{x}
}{
	\sqrt{1 + \frac{\pi}{8}\textrm{Var}\!\sq{x}}
}}
\label{eqn:probit}
\end{equation}%
\vspace*{-1mm}

With the above approximations,
we can calculate \cref{eqn:gen-pred} efficiently.
However, because the distribution over predicates in~\cref{eqn:gen-pred}
only depends on \textit{relative} probabilities of truth,
the model might learn to keep them all close to~0,
which would damage the logical interpretation of the model.
To avoid this, I have modified
the second term of \cref{eqn:grad}
and second term of \cref{eqn:grad-phi},
using not only the probability of
\textit{generating} a predicate for a pixie,~$\Pc{r}{x}$,
but also the probability of
the \textit{truth} of a predicate,~$\Pc{t_{r,X}}{x}$.
This technique of constraining latent variables
to improve interpretability is similar to
how \citet{rei2018attention} constrain attention weights.

Finally, as with other autoencoder models,
there is a danger of learning an identity function that generalises poorly.
Here, the problem is that the pixie distribution for a node
might be predicted based purely on the observed predicate for that node,
ignoring the wider context.
To avoid this problem, we can use \textit{dropout} on the input,
a technique which has been effective for other NLP models
\citep{iyyer2015textclass,bowman2016vae},
and which is closely related to denoising autoencoders \citep{vincent2008denoise}.
More precisely, we can keep the graph topology intact,
but randomly mask out the predicates for some nodes.
For a masked node~$X$,
I have initialised the encoder
with an embedding as shown in~\cref{eqn:dropout},
which depends on the node's dependencies
(only on the label of each dependency,
not on the predicate of the other node).
\vspace{-1mm}
\begin{equation}
e^{(X)}
= e^{(\textrm{drop})}
+ \!\!\sum_{Y\xleftarrow{l}X} \!\!\! e^{(\textrm{drop},l)} \\
+ \!\!\sum_{Y\xrightarrow{l}X} \!\!\! e^{(\textrm{drop},l^{-1})}
\label{eqn:dropout}
\end{equation}

\subsection{Approximating the Prior Expectation}
\label{sec:approach:bp}

The previous section explains
the inference network and how it is trained.
To train the generative model,
the predictions of the inference network (without dropout)
are used to approximate the conditional expectations~$\E{s|g}$ in~\cref{eqn:grad}.
However, the prior expectation~$\E{s}$
cannot be calculated using the inference network.
Intuitively, the prior distribution encodes a world model,
and this cannot be summarised as a single mean-field situation.

\citet{emerson2016} propose an MCMC algorithm using persistent particles,
summing over samples to approximate the expectation.
Many samples are required for a good approximation,
which is computationally expensive.
Taking a small number produces high variance gradients,
which makes training less stable.

However, we can see in~\cref{eqn:grad}
that we don't need the prior expectation~$\E{s}$ on its own,
but rather the difference ${\E{s|g}\mn\E{s}}$.
So, to reduce the variance of gradients,
we can try to explore the prior distribution
only in the vicinity of the inference network's predictions.
In particular, 
I propose taking the inference network's predictions
and updating this mean-field distribution
to bring it closer to the prior under the generative model.
This can be done using belief propagation
(for an introduction, see: \citealp{yedidia2003belief}),
as applied to CaRBMs by \citet{swersky2012carbm}.
For example, given the predicted mean-field vectors
for a gardener cutting grass,
we would modify these vectors to make the distribution
more closely match what is plausible under the generative model
(based on the world model, ignoring the observed predicates).

This can be seen as the bias-variance trade-off:
the inference network introduces a bias,
but reduces the variance, thereby making training more stable.

\subsection{Summary}
\label{sec:approach:summary}

The Pixie Autoencoder is a combination of the generative model
from Functional Distributional Semantics
(generating dependency graphs from latent situations)
and an inference network
(inferring latent situations from dependency graphs),
as illustrated in \cref{fig:graph,fig:encoder}.
They can be seen as an decoder and encoder, respectively.

It is trained on a sembank,
with the generative model maximising the likelihood of the dependency graphs,
and the inference network minimising KL-divergence with the generative model.
To calculate gradients,
the inference network is first applied to a dependency graph to infer the latent situation.
The generative model gives the energy of the situation
and the likelihood of the observed predicates
(compared with random predicates).
We also calculate the entropy of the situation,
and apply belief propagation to get a situation closer to the prior.
This gives us all terms in \cref{eqn:grad,eqn:grad-phi}.

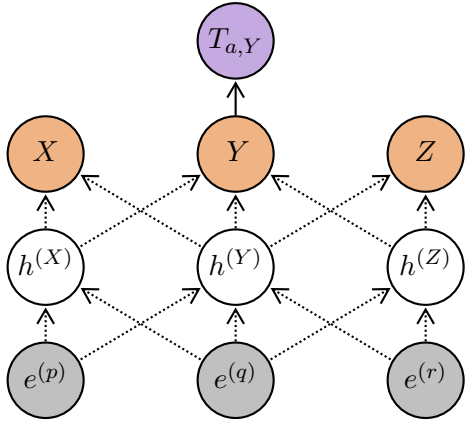
\begin{figure}
	\centering
	\begin{tikzpicture}[defshort]
	
	
	\node[ent] (y) {$Y\!$} ;
	\node[ent, right=of y] (z) {$Z$} ;
	\node[ent, left=of y] (x) {$X$} ;
	
	
	\node[node, below=of x] (tx) {$h^{(X)}$} ;
	\node[node, below=of y] (ty) {$h^{(Y)}$} ;
	\node[node, below=of z] (tz) {$h^{(Z)}$} ;
	
	
	\node[tok, below=of tx] (p) {$e^{(p)}$} ;
	\node[tok, below=of ty] (q) {$e^{(q)}$} ;
	\node[tok, below=of tz] (r) {$e^{(r)}$} ;
	
	
	\draw[inf] (tx) -- (p);
	\draw[inf] (ty) -- (p);
	\draw[inf] (tx) -- (q);
	\draw[inf] (ty) -- (q);
	\draw[inf] (tz) -- (q);
	\draw[inf] (ty) -- (r);
	\draw[inf] (tz) -- (r);
	\draw[inf] (x) -- (tx);
	\draw[inf] (y) -- (tx);
	\draw[inf] (x) -- (ty);
	\draw[inf] (y) -- (ty);
	\draw[inf] (z) -- (ty);
	\draw[inf] (y) -- (tz);
	\draw[inf] (z) -- (tz);
	
	
	\node[val, above=of y] (t) {$T_{a,Y}$};
	\draw[dir] (y) -- (t);
	
	\end{tikzpicture}%
	\caption{%
		An example of logical inference, building on \cref{fig:encoder}.
		Given an observed semantic dependency graph
		(here, with three nodes, like \cref{fig:dmrs},
		with predicates $p$,~$q$,~$r$),
		we would like to know if some predicate is true
		of some latent individual
		(here, if $a$ is true of~$Y$).
		We can apply the inference network
		to infer distributions for the pixie nodes,
		and then apply a semantic function to a pixie node
		(here, the function for~$a$ applied to~$Y$).
	}%
	\label{fig:inf}%
	\vspace*{-1mm}
\end{figure}

A strength of the Pixie Autoencoder
is that it supports logical inference,
following \citet{emerson2017a}.
This is illustrated in \cref{fig:inf}.
For example, for a gardener cutting grass or a child cutting a cake,
we could ask whether the cutting event is also a slicing event or a mowing event.

\subsection{Structural Prior}
\label{sec:approach:prior}

I have motivated the Pixie Autoencoder
from the perspective of the generative model.
However, we can also view it from the perspective of the encoder,
comparing it with a Variational Autoencoder (VAE)
which uses an RNN to generate text from a latent vector \citep{bowman2016vae}.
The VAE uses a Gaussian prior,
but the Pixie Autoencoder has a structured prior defined by the world model.

\citet{hoffman2016elbo} find that VAEs struggle to fit a Gaussian prior.
In contrast, the Pixie Autoencoder \textit{learns} the prior,
fitting the world model to the inference network's predictions.
Since the world model makes structural assumptions,
defining energy based only on semantic dependencies,
we can see the world model as a ``structural prior'':
the inference network is encouraged,
via the first term in~\cref{eqn:grad-phi},
to make predictions that can be modelled under these structural assumptions.

\section{Experiments and Evaluation}
\label{sec:eval}

I have evaluated on two datasets, chosen for two reasons.
Firstly, they allow a direct comparison with previous results \citep{emerson2017b}.
Secondly, they require fine-grained semantic understanding,
which starts to use the expressiveness of a functional model.

More open-ended tasks such as lexical substitution and question answering
would require combining my model with additional components
such as a semantic parser and a coreference resolver.
Robust parsers exist which are compatible with my model
\citeg{buys2017parse,chen2018parse},
but this would be a non-trivial extension,
particularly for incorporating robust coreference resolution,
which would ideally be done hand-in-hand with semantic analysis.
Incorporating fine-grained semantics into such tasks
is an exciting research direction, but beyond the scope of the current paper.

When reporting results, significance tests follow \citet{dror2018significance}.

\subsection{Training Details}

I trained the model on WikiWoods
\citep{flickinger2010wikiwoods,solberg2012wikiwoods},
which provides DMRS graphs \citep{copestake2005mrs,copestake2009dmrs}
for 55m sentences (900m tokens)
from the English Wikipedia (July 2008).
It was parsed with the English Resource Grammar (ERG)
\citep{flickinger2000erg,flickinger2011erg}
and PET parser \citep{callmeier2001pet,toutanova2005pet},
with parse ranking trained on
WeScience \citep{ytrestol2009wescience}.
It is updated with each ERG release;
I used the 1212 version.
I preprocessed the data following \citet{emerson2016},
giving 31m graphs.

I implemented the model using DyNet \citep{neubig2017dynet}
and Pydmrs \citep{copestake2016pydmrs}.\footnote{%
	\url{https://gitlab.com/guyemerson/pixie}
}
I initialised the generative model
following \citet{emerson2017b}
using sparse PPMI vectors \citep{qasemizadeh2016vector}.
I first trained the encoder on the initial generative model,
then trained both together.
I used L2 regularisation
and the Adam optimiser \citep{kingma2015adam},
with separate L2 weights and learning rates
for the world model, lexical model, and encoder.
I tuned hyper\-parameters on the RELPRON dev set (see~\cref{sec:eval:relpron}),
and averaged over 5 random seeds.

\subsection{BERT Baseline}
\label{sec:eval:bert}

BERT \citep{devlin2019bert} is a large pre-trained language model
with a Transformer architecture \citep{vaswani2017attention},
trained on 3.3b tokens
from the English Wikipedia and BookCorpus \citep{zhu2015books}.
It produces high-quality contextualised embeddings,
but its architecture is not motivated by linguistic theory.
I used the version in the Transformers library \citep{wolf2019transformer}.
To my knowledge,
large language models have not previously been evaluated on these datasets.

\subsection{RELPRON}
\label{sec:eval:relpron}

\begin{table*}[t]
	\centering
	\begin{tabular}{|l|l|c|c|}
		\cline{2-4}
		\multicolumn{1}{l|}{}
		& Model & Dev & Test \\
		\hline
		\multirow{6}{*}{Previous work}
		& Vector addition \citep{rimell2016relpron} & .496 & .472 \\
		& Simplified Practical Lexical Function \citep{rimell2016relpron} & .496 & .497 \\
		& Vector addition \citep{czarnowska2019} & .485 & .475 \\
		& Dependency vector addition \citep{czarnowska2019} & .497 & .439 \\
		\cline{2-4}
		& Semantic functions \citep{emerson2017b} & .20\hphantom{0} & .16\hphantom{0} \\
		& Sem-func \& vector ensemble \citep{emerson2017b} & .53\hphantom{0} & .49\hphantom{0} \\
		\hline
		\multirow{4}{*}{Baselines}
		& Vector addition & .488 & .474 \\
		& BERT (masked prediction) & .206 & .186 \\
		& BERT (contextual prediction) & .093 & .134 \\
		& BERT (masked prediction) \& vector addition ensemble & .498 & .479 \\
		\hline
		\multirow{2}{*}{Proposed approach}
		& Pixie Autoencoder & .261 & .189 \\
		& Pixie Autoencoder \& vector addition ensemble & .532 & .489 \\
		\hline
	\end{tabular}
	\vspace*{-2mm}
	\caption{%
		Mean Average Precision (MAP) on RELPRON development and test sets.
	}
	\vspace*{-2mm}
	\label{tab:relpron}
\end{table*}

The RELPRON dataset \citep{rimell2016relpron}
consists of \textit{terms} (such as \textit{telescope}),
paired with up to 10 \textit{properties} (such as \textit{device that astronomer use}).
The task is to find the correct properties for each term.
There is large gap between the state of the art (around 50\%)
and the human ceiling (near 100\%).

The dev set contains 65 terms and 518 properties;
the test set, 73 terms and 569 properties.
The dataset is too small to train on,
but hyperparameters can be tuned on the dev set.
The dev and test terms are disjoint,
to avoid high scores from overtuning.

Previous work has shown that vector addition performs well on this task \citep{rimell2016relpron,czarnowska2019}.
I have trained a Skip-gram model \citep{mikolov2013vector}
using the Gensim library \citep{rehurek2010gensim},
tuning weighted addition on the dev set.

For the Pixie Autoencoder, we can view the task as logical inference,
finding the probability of truth of a term given an observed property.
This follows \cref{fig:inf},
applying the term~$a$ to either $X$ or~$Z$,
according to whether the property has a subject or object relative clause.

BERT does not have a logical structure,
so there are multiple ways we could apply it.
I explored many options,
to make it as competitive as possible.
Following \citet{petroni2019knowledge},
we can rephrase each property as a cloze sentence
(such as \textit{a device that an astronomer uses is a [MASK]~.}).
However, RELPRON consists of pseudo-logical forms,
which must be converted into plain text query strings.
For each property, there are many possible cloze sentences,
which yield different predictions.
Choices include:
grammatical number,
articles,
relative pronoun,
passivisation,
and position of the mask.
I used the Pattern library \citep{smedt2012pattern}
to inflect words for number.

Results are given in \cref{tab:relpron}.
The best performing BERT method
uses singular nouns with \textit{a}/\textit{an},
despite sometimes being ungrammatical.
My most careful approach involves
manually choosing articles
(e.g.\ \textit{a device}, \textit{the sky}, \textit{water})
and number (e.g.\ plural \textit{people})
and trying three articles for the masked term
(\textit{a}, \textit{an}, or no article,
taking the highest probability from the three),
but this actually lowers dev set performance to {.192}.
Using plurals lowers performance to {.089}.
Surprisingly, using BERT large (instead of BERT base) lowers performance to {.165}.
As an alternative to cloze sentences,
BERT can be used to predict the term from a contextualised embedding.
This performs worse (see \cref{tab:relpron}),
but the best type of query string is similar.

The Pixie Autoencoder outperforms
previous work using semantic functions,
but is still outperformed by vector addition.
Combining it with vector addition in a weighted ensemble
lets us test whether they have learnt different kinds of information.
The ensemble significantly outperforms vector addition
on the test set
($p<0.01$ for a permutation test),
while the BERT ensemble does not
($p>0.2$).
However, it performs no better than the ensemble in previous work.
This suggests that, while the encoder has enabled
the model to learn more information,
the additional information is already present in the vector space model.

RELPRON also includes a number of \textit{confounders},
properties that are challenging due to lexical overlap.
For example, an \textit{activity that soil supports}
is \textit{farming}, not \textit{soil}.
There are 27 confounders in the test set,
and my vector addition model places all of them
in the top 4 ranks for the confounding term.
In contrast, the Pixie Autoencoder and BERT do not fall for the confounders,
with a mean rank of 171 and 266, respectively.

Nonetheless, vector addition remains hard to beat.
As vector space models are known to be good at topical relatedness
(e.g.\ learning that \textit{astronomer} and \textit{telescope}
are related, without necessarily learning how they are related),
a tentative conclusion is that relatedness is missing
from the contextualised models (Pixie Autoencoder and BERT).
Finding a principled way to integrate a notion of ``topic''
would be an interesting task for future work.

\begin{table*}[t]
	\centering
	\begin{tabular}{|c|l|c|c|}
		\cline{2-4}
		\multicolumn{1}{l|}{}
		& Model & Separate & Averaged \\
		\hline
		\multirow{8}{*}{\shortstack{Previous\\ work}}
		& Vector addition \citep{milajevs2014tensor} & - & .348 \\
		& Categorical, copy object \citep{milajevs2014tensor} & - & .456 \\
		& Categorical, regression \citep{polajnar2015sentence} & .33\hphantom{0} & - \\
		& Categorical, low-rank decomposition \citep{fried2015rank} & .34\hphantom{0} & - \\
		& Tensor factorisation \citep{vandecruys2013tensor} & .37\hphantom{0} & - \\
		& Neural categorical \citep{hashimoto2014neural} & .41\hphantom{0} & .50\hphantom{0} \\
		\cline{2-4}
		& Semantic functions \citep{emerson2017b} & .25\hphantom{0} & - \\
		& Sem-func \& vector ensemble \citep{emerson2017b} & .32\hphantom{0} & - \\
		\hline
		\multirow{2}{*}{Baselines}
		& BERT (contextual similarity) & .337 & .446 \\
		& BERT (contextual prediction) & .233 & .317 \\
		\hline
		\multirow{2}{*}{\shortstack{Proposed\\[-1mm]approach\vspace*{-1mm}}}
		& Pixie Autoencoder (logical inference in both directions) & .306 & .374 \\
		& Pixie Autoencoder (logical inference in one direction) & .406 & .504 \\
		\hline
	\end{tabular}
	\vspace*{-2mm}
	\caption{%
		Spearman rank correlation on the GS2011 dataset,
		using separate or averaged annotator scores.
	}
	\vspace*{-2mm}
	\label{tab:gs2011}
\end{table*}

\subsection{GS2011}
\label{sec:eval:gs2011}

The GS2011 dataset evaluates similarity in context
\citep{grefenstette2011svo}.
It comprises pairs of verbs combined with the same subject and object
(for example, \textit{map show location} and \textit{map express location}),
annotated with similarity judgements.
There are 199 distinct pairs, and 2500 judgements (from multiple annotators).

Care must be taken when considering previous work, for two reasons.
Firstly, there is no development set.
Tuning hyperparameters directly on this dataset will lead to artificially high scores,
so previous work cannot always be taken at face value.
For example, \citet{hashimoto2014neural} report results for 10 settings.
I nonetheless show the best result in \cref{tab:gs2011}.
My model is tuned on RELPRON (\cref{sec:eval:relpron}).

Secondly, there are two ways to calculate correlation with human judgements:
averaging for each distinct pair,
or keeping each judgement separate.
Both methods have been used in previous work,
and only \citet{hashimoto2014neural} report both.

For the Pixie Autoencoder, we can view the task as logical inference,
following \cref{fig:inf}.
However, \citet{vandecruys2013tensor} point out that
the second verb in each pair
is often nonsensical when combined with the two arguments
(e.g.\ \textit{system visit criterion}),
and so they argue that only the first verb should be contextualised,
and then compared with the second verb.
This suggests we should
apply logical inference only in one direction:
we should find the probability of truth of the second verb,
given the first verb and its arguments.
As shown in \cref{tab:gs2011},
this gives better results than
applying logical inference in both directions
and averaging the probabilities.
Logical inference in both directions
allows a direct comparison with \citet{emerson2017b},
showing the Pixie Autoencoder performs better.
Logical inference in one direction
yields state-of-the-art results on par with
the best results of \citet{hashimoto2014neural}.

There are multiple ways to apply BERT, as in~\cref{sec:eval:relpron}.
One option is to calculate cosine similarity of contextualised embeddings
(averaging if tokenised into word-parts).
However, each subject-verb-object triple must be converted to plain text.
Without a dev set,
it is reassuring that conclusions from RELPRON carry over:
it is best to use singular nouns with \textit{a}/\textit{an}
(even if ungrammatical)
and it is best to use BERT base.
Manually choosing articles and number
lowers performance to {.320} (separate), 
plural nouns to {.175},
and BERT large to {.226}.
Instead of using cosine similarity,
we can predict the other verb from the contextualised embedding,
but this performs worse.
The Pixie Autoencoder outperforms BERT,
significantly for separate scores
($p<0.01$ for a bootstrap test),
but only suggestively for averaged scores
($p=0.18$).

\section{Conclusion}

I have presented the Pixie Autoencoder,
a novel encoder architecture and training algorithm
for Functional Distributional Semantics,
improving on previous results in this framework.
For GS2011, the Pixie Autoencoder achieves state-of-the-art results.
For RELPRON, it learns information
not captured by a vector space model.
For both datasets, it outperforms BERT,
despite being a shallower model with fewer parameters,
trained on less data.
This points to the usefulness
of building semantic structure into the model.
It is also easy to apply to these datasets
(with no need to tune query strings),
as it has a clear logical interpretation.

\section*{Acknowledgements}

Above all, I have to thank my PhD supervisor Ann Copestake,
without whose guidance my work would be far less coherent.
I would like to thank my PhD examiners
Katrin Erk and Paula Buttery,
for discussions which helped to clarify the next steps I should work on.
I would like to thank the many people who I have discussed this work with,
particularly Kris Cao, Amandla Mabona, and Lingpeng Kong for advice on training VAEs.
I would also like to thank Marek Rei, Georgi Karadzhov,
and the NLIP reading group in Cambridge,
for giving feedback on an early draft.
Finally, I would like to thank the anonymous ACL reviewers,
for pointing out sections of the paper that were unclear
and suggesting improvements.

I am supported by a Research Fellowship at Gonville \& Caius College, Cambridge.

\bibliography{thesis,pixie}
\bibliographystyle{acl_natbib}


\end{document}